\documentclass[11pt, a4paper, logo, twocolumn, copyright]{googledeepmind}

\usepackage[authoryear, sort&compress, round]{natbib}
\usepackage[textwidth=18mm]{todonotes}
\usepackage{graphicx}
\usepackage{makecell}
\usepackage{amsmath}
\usepackage{multirow}
\usepackage{caption}
\usepackage{etoolbox}
\usepackage[bottom]{footmisc}

\def\by{{\mathbf{y}}}

\title{Adapting Decoder-Based Language Models for Diverse Encoder Downstream Tasks}

\author{Paul Suganthan*, Fedor Moiseev, Le Yan, Junru Wu, Jianmo Ni, Jay Han, Imed Zitouni, Enrique Alfonseca, Xuanhui Wang, Zhe Dong* \\
\{paulgc, zhedong\}@google.com, Google Inc.}

\begin{abstract}
Decoder-based transformers, while revolutionizing language modeling and scaling to immense sizes, have not completely overtaken encoder-heavy architectures in natural language processing.  Specifically, encoder-only models remain dominant in tasks like classification, regression, and ranking. This is primarily due to the inherent structure of decoder-based models, which limits their direct applicability to these tasks. In this paper, we introduce \textit{Gemma Encoder}, adapting the powerful Gemma decoder model to an encoder architecture, thereby unlocking its potential for a wider range of non-generative applications. To optimize the adaptation from decoder to encoder, we systematically analyze various pooling strategies, attention mechanisms, and hyperparameters (e.g., dropout rate). Furthermore, we benchmark Gemma Encoder against established approaches on the GLUE benchmarks, and MS MARCO ranking benchmark, demonstrating its effectiveness and versatility.
\end{abstract}

\begin{document}

\maketitle

\section{Introduction}

Decoder-based language models like Gemma~\citep{gemmateam2024gemmaopenmodelsbased, gemmateam2024gemma2improvingopen} and Gemini~\citep{geminiteam2024gemini} have demonstrated remarkable language understanding capabilities. Yet, for many downstream tasks such as classification, regression, and ranking, encoder-based models, particularly those derived from BERT~\citep{devlin-etal-2019-bert} or T5's encoder~\citep{raffel2020t5}, remain the dominant choice.  A key question thus arises: can we effectively adapt the powerful knowledge embedded in decoder-only models to excel in these encoder-centric tasks?

This work addresses this gap by introducing Gemma Encoder, a novel adaptation of the Gemma decoder model designed for encoder-only architectures.  We leverage Gemma's pre-trained weights as a strong initialization point, and then strategically modify the architecture and training procedure to optimize performance on downstream tasks. Our approach centers on three key innovations:

First, we augment the model with task-specific pooling and Multi-Layer Perceptron (MLP) layers, exploring various pooling strategies to determine the optimal architecture.

Second, we address the critical impact of attention mechanisms.  Gemma's causal attention, ideal for generative tasks, inherently limits its applicability to encoder-based tasks. We demonstrate that simply enabling bidirectional attention during fine-tuning dramatically improves performance.

Third, we investigate the role of dropout.  While often omitted during pre-training of modern decoder models, our empirical analysis reveals that incorporating dropout during fine-tuning significantly enhances Gemma Encoder's robustness and generalization ability. We also analyze different padding strategies to understand the effect on Encoder models.

In the rest of this paper, we describe the technical details of these modifications\footnote{A companion code release will be included with a forthcoming revision.}. To validate the effectiveness of our Gemma Encoder, we conduct our experiments on GLUE benchmarks~\citep{wang2019gluemultitaskbenchmarkanalysis, wang2020supergluestickierbenchmarkgeneralpurpose}, for classification and regression tasks, and the MSMARCO benchmark~\citep{bajaj2018msmarcohumangenerated} for ranking tasks. Our results show that our Gemma Encoder models are able to outperform competitive baselines on these benchmark tasks.

\begin{figure*}[h]
\centering
\includegraphics[width=.9\textwidth]{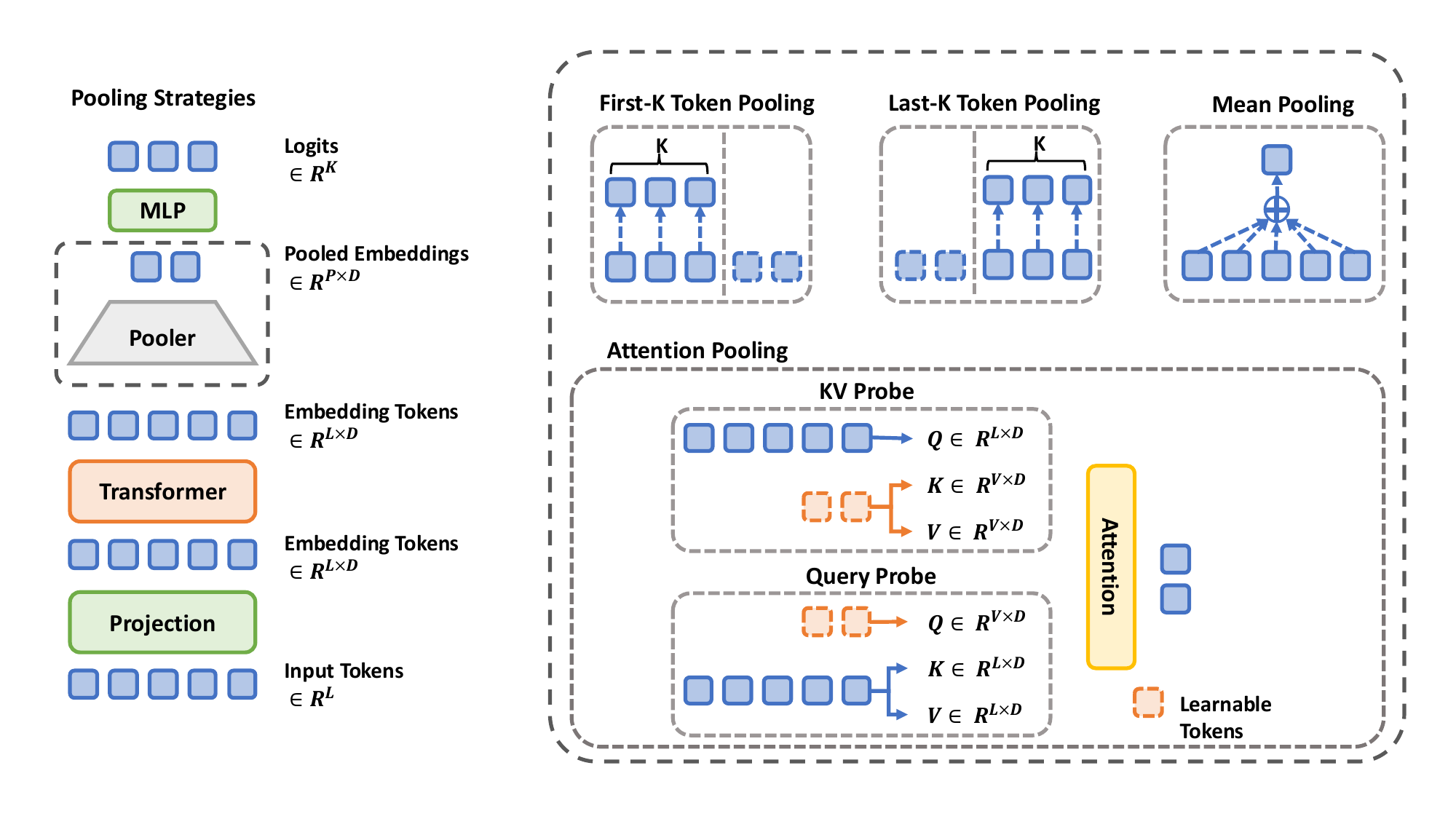}
\captionsetup{width=0.95\textwidth,font=footnotesize,justification=centering}
\caption{Gemma Encoder Architecture \& Pooling: The architecture (left) comprises an encoder transformer initialized from Gemma, followed by task-specific pooler and MLP layers. The right panel illustrates the various pooling strategies considered: \textbf{First-K}, \textbf{Last-K}, \textbf{Mean}, and \textbf{Attention Pooling} with \textit{KV-} and \textit{Query-probe} variants).} \label{gemma_fig}
\end{figure*}

\section{Model Adaptation Choices}
This work explores adapting decoder-only generative AI models, exemplified by GPT-4~\citep{openai2024gpt4technicalreport}, Gemini~\citep{geminiteam2024gemini}, and Gemma~\citep{gemmateam2024gemmaopenmodelsbased, gemmateam2024gemma2improvingopen}, for encoder-only tasks such as classification, regression, and ranking. Our adaptation strategy, as shown in Figure~\ref{gemma_fig}, centers on three key architectural and training modifications: attention mechanism design, pooling strategies, and the application of dropout.

While both decoder-only and encoder-only Transformers share a similar underlying layer design, their training objectives and downstream applications diverge significantly. Decoder-only models are optimized for next-token prediction, making them well-suited for generative tasks. In contrast, encoder-only models are trained to generate a comprehensive representation of the entire input sequence, empowering a diverse range of predictive tasks, including classification (label prediction), regression (score prediction), and ranking (relative ordering prediction).

By systematically evaluating these crucial modeling choices, we aim to identify the architecture that best captures the essential information embedded within the input sequence. This will ultimately optimize the performance of Gemma Encoder on downstream tasks. Our analysis provides valuable insights into the trade-offs inherent in various design decisions, offering guidance not only for adapting Gemma, but also for transforming other decoder-only models into effective encoders.

\subsection{Attention Masking}

Pre-trained generative models often employ three types of attention mask patterns: bidirectional, causal, and prefix masking (Figures 3 and 4 in~\citet{raffel2020t5}). For our focus on non-generative tasks, we limited the explorations to causal attention and bidirectional attention.
 
Bidirectional masking, also referred as fully-visible masking~\citep{raffel2020t5}, is commonly used in encoder models. It allows the encoder to generate a holistic representation of the input by providing complete access to all input tokens, fostering a comprehensive understanding of the entire sequence.

Causal masking, on the other hand, is prevalent in decoder-only and sequence-to-sequence models. Here, tokens are processed sequentially, and predictions for the next token rely solely on preceding tokens. This prevents the model from "looking ahead" during training, preserving the auto-regressive property essential for text generation.  The attention mechanism is masked so that each token attends only to itself and prior tokens.  

T5 introduced PrefixLM, a hybrid approach that utilizes causal masking with a designated "prefix" section. This prefix is processed bidirectionally, allowing the model to attend to all tokens within it. The remaining sequence is processed causally, enabling generation conditioned on the fully contextualized prefix. This combines the benefits of bidirectional context for understanding the initial input segment with the autoregressive capabilities of causal masking for generating the subsequent sequence.

Given that Gemma models are pre-trained with causal attention, we investigated the impact of both bidirectional and causal attention masks \textit{during fine-tuning} to maximize the performance of Gemma Encoder models. 

\subsection{Pooling Strategies}

The Gemma Encoder model aims to create a robust representation by effectively leveraging all input information. It comprises an encoder transformer initialized from the Gemma decoder transformer, and a pooler coupled with Multi-Layer Perceptron (MLP) layers. The pooling and MLP layers are randomly initialized. A crucial component for aggregating the contextualized token representations from the transformer into a fixed-length vector, we explored several pooling options to determine the optimal architecture.

\paragraph{First-K and Last-K token poolings} are the two simplest pooling strategies used in transformer models to extract a fixed-size representation from a sequence of tokens for encoder tasks. In \textit{First-K token pooling}, the representation of the first K tokens is used to aggregate information from the entire input sequence through attention mechanisms, with $K$ being the number of classes for classification or $1$ for regression and ranking. This approach can be viewed as fine-tuning the first K tokens as special tokens, analogous to the \texttt{[CLS]} token in BERT. In contrast, \textit{Last-K token pooling} takes the representation of the last token in the sequence. This approach is particularly relevant in scenarios like language modeling or when the final tokens represents a meaningful conclusion of the sequence, such as an \texttt{[EOS]} token. While \textit{both} approaches simplify the transformation of variable-length sequences into fixed-size representations, they may capture slightly different aspects of the input depending on the token's role in the model architecture and the training objective.

\paragraph{Mean pooling} is another parameter-free pooling method averages the hidden states across all tokens, creating a single vector representing the average contextualized information. Mean pooling is computationally efficient and provides a basic representation of the entire input. However, it can be susceptible to noise and may dilute the importance of critical tokens by treating all tokens equally. 

\paragraph{Attention pooling} is a more sophisticated technique that employs an attention mechanism to weight and aggregate token representations. By learning attention weights, the model can focus on the most informative tokens for the downstream task, effectively filtering noise and highlighting relevant information. Attention pooling offers greater flexibility and expressiveness compared to mean pooling. 

There are \textit{two main} approaches to implementing attention pooling. The attention mechanism can be expressed as:
\begin{equation*}
\mathrm{Output} = \mathrm{Softmax}(\frac{Q \cdot K^\mathrm{T}}{\sqrt{D}}) \cdot V,
\end{equation*}
where the input is passed through the query matrix, $ Q \in \mathbb{R}^{L \times D}$, and the latent variables, $K, V \in \mathbb{R}^{V \times D}$, are learned. Here, $L$, $D$, and $V$ represent the input sequence length, the internal embedding dimension, and the number of latent variables, respectively. This approach is represented as \textbf{KV-probe} in Figure~\ref{gemma_fig}.

Alternatively, the input can be passed through the key and value matrices, $K, V \in \mathbb{R}^{L \times D}$, while the latent variables are learned through the query matrix, $Q \in \mathbb{R}^{V \times D}$, as introduced in~\citet{jaegle2022perceiver}. This approach is represented as \textbf{Query-probe} in Figure~\ref{gemma_fig}.

\begin{table*}
\small
\centering
\begin{tabular}{c c c c c c}
  \hline
  Benchmark & Dataset  & Metric & \#Train & \#Eval & Task \\ 
  \hline
  \multirow{8}{*}{GLUE} 
  & STSB & Spearman Coeff. & 5749 & 1500 & \texttt{Similarity} \\ 
  & CoLA & Matthews Coeff. & 8551 & 1043 & \texttt{Acceptability} \\ 
  & QQP & F1 & 363846 & 40430 & \texttt{Paraphrase} \\ 
  & QNLI & Accuracy & 104743 & 5463 & \texttt{QA} / \texttt{NLI}\\ 
  & SST2 & Accuracy & 67349 & 872 & \texttt{Sentiment} \\ 
  & RTE & Accuracy & 2490 & 277 &  \texttt{NLI} \\ 
  & MRPC & F1 & 3668 & 408 & \texttt{Paraphrase}\\ 
  & MNLI-matched & Accuracy & 392702 & 9815 & \texttt{NLI} \\ 
  & MNLI-mismatched & Accuracy & 392702 & 9832 & \texttt{NLI} \\ 
  \hline
  \multirow{7}{*}{SuperGLUE} & BoolQ & Accuracy & 9427 & 3270 & \texttt{QA} \\ 
  & RTE & Accuracy & 2490 & 277 & \texttt{NLI} \\ 
  & COPA & Accuracy & 400 & 100 & \texttt{QA} \\ 
  & CB & Accuracy & 250 & 56 & \texttt{NLI} \\ 
  & WIC & Accuracy & 5428 & 638 & \texttt{WSD}\\ 
  & WSC & Accuracy & 554 & 104 & \texttt{Resolution} \\ 
  & MULTIRC & F1 & 27243 & 4848 & \texttt{QA} \\
  \hline
  Ranking & MS-MARCO & MRR/NDCG & 532751 & 6980 & \texttt{Ranking} \\
  \hline
\end{tabular}
\captionsetup{width=0.95\textwidth,font=footnotesize,justification=centering}
\caption{Dataset Statistics across GLUE, SuperGLUE and Ranking Benchmarks used in the paper. The benchmarks cover a diverse range of encoder tasks, including sentence similarity (\texttt{similarity}), \texttt{acceptability}, \texttt{paraphrase}, question answering (\texttt{QA}), natural language inference (\texttt{NLI}), sentiment classification (\texttt{sentiment}), word sense disambiguation (\texttt{WSD}), coreference resolution (\texttt{resolution}), and document ranking (\texttt{ranking}.) \label{data-stats}}
\end{table*}

\subsection{Dropout}

Dropout is a crucial regularization technique widely used in deep learning to prevent over-fitting and improve generalization by randomly deactivating a fraction of neurons during training. However, overfitting is less of a concern for the latest large language models (LLMs), due to the massive training datasets and high model capacity, which naturally provide robust generalization. Instead of dropout, LLMs~\citep{chowdhery2022palmscalinglanguagemodeling, anil2023palm2technicalreport, hoffmann2022trainingcomputeoptimallargelanguage} rely on other forms of regularization, such as weight decay or careful scaling strategies, which are better suited to the massive scale of these models and training corpora.

When adapting a decoder-only model to an encoder-only architecture, evaluating the role of dropout through ablation studies is crucial. Encoder-only tasks, such as regression, classification, and ranking, are more prone to overfitting compared to generative tasks handled by decoder-only models. This increased susceptibility arises because encoder-only models are typically trained on limited data using supervised fine-tuning, with less diversity in the output space. Moreover, the supervision signal in encoder-only tasks operates at the sentence or full-input-sequence level, relying on deterministic loss functions. This setup makes the model more likely to fit spurious features in the input. In contrast, autoregressive tasks in decoder-only models inherently involve uncertainty and softer probability distributions, which naturally help mitigate overfitting.

\begin{table*}[h]
\small
\centering
\begin{tabular}{c c | c c c c c c c c c c}
  \hline
  \multirow{2}{*}{Model} & \multirow{2}{*}{Size} & \multirow{2}{*}{STSB} & \multirow{2}{*}{CoLA} & \multirow{2}{*}{QQP} & \multirow{2}{*}{QNLI} & \multirow{2}{*}{SST2} & \multirow{2}{*}{RTE} & \multirow{2}{*}{MRPC} & \multicolumn{2}{c}{MNLI} & \multirow{2}{*}{Avg}\\
  &&&&&&&&& m & mm &\\
 \hline
    BERT-base & 108M & 84.6 & 56.9 & 87.8 & 90.4 & 92.2 & 70.0 & 90.6 & 83.1 & 83.7 & 82.1 \\ 

    T5-base & 110M & 85.2 & 53.4 & 89.5 & 93.2 & 94.7 & 69.7 & 91.6 & 88.8 & 88.4 & 83.8 \\ 

    BERT-large & 334M & 86.4 & 63.9 & 88.4 & 93.0 & 94.0 & 75.5 & 92.4 & 87.1 & 86.9 & 85.3 \\ 

    T5-large & 350M & 88.0 & 63.6 & 89.6 & 94.8 & 96.9 & 85.6 & 93.5 & 91.0 & 90.9 & 88.2 \\ 

    T5-xl & 1.5B & 87.4 & 70.9 & 90.3 & 96.2 & 97.0 & 92.1 & 93.5 & 92.1 & 91.7 & 90.1 \\ 
  
    T5-xxl & 6.5B & 86.9 & \textbf{72.9} & \textbf{90.4} & \textbf{96.4} & \textbf{97.2} & 92.8 & \textbf{94.2} & 92.1 & 92.0 & 90.5 \\ 
\hline

    Gemma-2 2B & 2B & 92.4 & 67.7 & 89.2 & 95.4 & 97.0 & 87.0 & 91.6 & 91.1 & 91.1 & 89.2 \\ 

    Gemma-2 9B & 9B & \textbf{92.6} & 71.3 & 90.1 & \textbf{96.4} & 96.7 & \textbf{93.1} & 93.9 & \textbf{92.2} & \textbf{92.1} & \textbf{90.9} \\ 
  \hline
\end{tabular}
\captionsetup{width=0.9\textwidth,font=footnotesize,justification=centering}
\caption{Evaluation on GLUE benchmarks. For MNLI task, \textbf{m} and \textbf{mm} refers to \texttt{matched} and \texttt{mismatched} accuracy. All the Gemma experiments are done with bidirectional attention masking, dropout on attention softmax and feedforward network outputs with 10\% rate, and right padding. The metrics for Gemma 2B and 9B are based on the best pooling strategies as ablated in Table~\ref{pooling-summary}.  \label{glue-results}}
\end{table*}

\begin{table*}[h]
\small
\centering
\begin{tabular}{c c | c c c c c c c c}
  \hline
  Model & Size & BoolQ & RTE & COPA & CB & WIC & WSC & MULTIRC & Avg\\ 
  \hline
  T5-xxl & 6.5B & 90.8 & \textbf{93.9} & \textbf{98.0} & 96.4 & \textbf{77.3} & \textbf{96.2} & 87.4 & \textbf{91.4} \\
 
  Gemma-2 2B & 2B & 88.8 & 88.8 & 90.0 & \textbf{100.0} & 75.1 & 73.1 & 85.6 & 85.9 \\ 
  Gemma-2 9B & 9B & \textbf{91.3} & 93.1 & 96.0 & \textbf{100.0} & 76.5 & 90.4 & \textbf{88.2} & 90.8 \\ 
  \hline
\end{tabular}
\captionsetup{width=0.95\textwidth,font=footnotesize,justification=centering}
\caption{Evaluation on SuperGLUE benchmarks. The same hyper-parameter settings are applied as in Table~\ref{glue-results}. \label{superglue-results}}
\end{table*}

\subsection{Padding Strategies}

When processing sequences of varying lengths in batched inputs, padding is essential to create uniform input tensors. However, the choice between \textbf{left padding} (prepending padding tokens) and \textbf{right padding} (appending padding tokens) may impacts model behavior, especially when adapting encoder model from decoder-only models.

In decoder-only transformer models, such as those used for language generation (e.g., GPT~\citep{openai2024gpt4technicalreport},  Gemma~\citep{gemmateam2024gemma2improvingopen, gemmateam2024gemmaopenmodelsbased} models), left padding is typically used to align the inputs during training or inference, due to the auto-regressive training objective and efficient positional embedding. Decoder-only models are trained in an auto-regressive manner, where the task is to predict the next token based on all previous tokens. Decoder-only transformer models use positional embeddings to encode the order of tokens. Left-padding ensures that the relative positions of the actual tokens remain consistent regardless of the sequence length.

However, right padding is acceptable for encoder models because of the way these models process input sequences. Unlike decoder-only models, encoders like those in BERT~\citep{devlin-etal-2019-bert}, T5~\citep{raffel2020t5} or other bidirectional transformer models handle the entire input sequence at once, and their attention mechanisms allow tokens to attend to any other token in the input through bidirectional attention. In encoder models, positional embeddings are applied to the entire sequence, including padding tokens. Since the padding tokens are ignored during attention, their positional embeddings don't interfere with the actual computation. Whether the padding is on the right or left does not affect the functionality.

The choice of padding strategy has implications for the pooling layer, especially in conjunction with causal attention and First-K/Last-K token pooling.

\subsection{Adaption for Ranking Tasks}
A ranking task concerns the relative ordering of a set of text documents by their relevance to a given query. Formally, for each query $q_i$, we are provided with a list of candidate documents $D_i = (d_{i1}, ..., d_{im})$ along with their relevance labels $y_i = (y_{i1}, ..., y_{im})$. The objective is to train a ranking model $f$, such that the model takes a query-document pair, $(q_i, d_{ij})$, as input and outputs a ranking score $\hat{y}_{ij} = f(q_i, d_{ij}) \in R$.

\paragraph{Listwise Inputs and Outputs.}
Adapting the Gemma Encoder to ranking tasks requires handling \textbf{listwise} inputs, which have the shape $[B, M, L]$, where $B$ represents the input batch size, $M$ the number of documents per query, and $L$ the sequence length.  To process these inputs, we first flatten the listwise structure to a shape of $[B\times M, L]$. This flattened input is then passed through the Gemma Encoder, resulting in logits with a shape of $[B\times M]$. Finally, these logits are reshaped back to $[B, M]$ to align with the expected input format of the ranking loss function.

\paragraph{Ranking Losses.}
This adaptation enables the model to produce a ranking score for each document. Consequently, any established ranking loss functions can be applied, such as pairwise logistic loss~\citep{burges2005learning}, PolyLoss~\citep{leng2022polyloss}, and Gumbel approximated NDCG loss~\citep{gumbel2020}. In this study, we focused on the \textit{listwise softmax cross-entropy} loss~\citep{bruch2019softmax, liu2011learning}.

\begin{equation*}
    \ell_{\text{Softmax}}(\by_i, \hat{\by}_i) = -\sum_{j=1}^{m} y_{ij} \log\Big(\frac{e^{\hat{y}_{ij}}}{\sum_{j'} e^{\hat{y}_{ij'}}}\Big).
\end{equation*}

\section{Experiments}

We evaluate the effectiveness of various modeling choices for adapting the decoder-only generative model, Gemma, to encoder-only tasks.  Performance is assessed across a diverse set of classification, regression, and ranking tasks. 

\subsection{Data}

Specifically, we demonstrate the effectiveness of Gemma Encoder for classification and scoring tasks on the GLUE~\citep{wang2019gluemultitaskbenchmarkanalysis} and SuperGLUE~\citep{wang2020supergluestickierbenchmarkgeneralpurpose} benchmarks. We further evaluate its ranking capability on the MSMARCO benchmark~\citep{bajaj2018msmarcohumangenerated}.

Gemma Encoder's general language understanding abilities were evaluated on the GLUE and SuperGLUE benchmarks, which include diverse text classification and scoring tasks. The tasks and their corresponding evaluation metrics are summarized in Table~\ref{data-stats}. Our evaluation involved finetuning Gemma Encoder on each task's training set and subsequently assessing its performance on the evaluation set. The RECORD task from SuperGLUE was excluded due to its incompatibility with an encoder-only architecture.

\subsection{Model}
We evaluated the encoder adaptation using the Gemma-2~\citep{gemmateam2024gemma2improvingopen}, specifically, the 2B and 9B decoder model variants.

\subsection{Evaluation}
\paragraph{GLUE and SuperGLUE evaluation}
We evaluate Gemma Encoder performance on GLUE, in Table~\ref{glue-results}, and SuperGLUE, in Table~\ref{superglue-results}, benchmarks and compare it with baseline T5 and BERT models. Notably, Gemma Encoder achieves competitive performance against similarly sized T5 models, despite not utilizing any uptraining or encoder-decoder masked language modeling (MLM) pretraining. This highlights the efficacy of our architecture adaptation approach in unlocking the potential of decoder-based language models for encoder-based tasks. 

The results presented in Tables~\ref{glue-results} and~\ref{superglue-results} reflect the optimal hyperparameter configuration identified through ablation studies (Section~\ref{sec:ablation}). Specifically, we found that bidirectional attention masking (Section~\ref{sec:attention_ablation}), right padding (Section~\ref{sec:padding_ablation}), and a 10\% dropout rate applied to the attention softmax and feedforward network outputs (Section~\ref{sec:dropout_ablation}) yielded the best performance.

Using the described training configurations, \textit{especially} with bidirectional attention, no single pooling strategy consistently optimized performance for both Gemma 2B and 9B. The choice of pooling strategy had negligible impact when using our finetuning dataset. However, \textit{with} causal attention masking (Gemma's default decoder setting), last-token pooling significantly outperformed other strategies during finetuning. Please refer to \ref{sec:pooling_ablation} for more details.

\paragraph{MS MARCO Ranking Evaluation} For the ranking task, we utilized the MS MARCO dataset, which contains approximately $530K$ queries in the \texttt{train} partition and $7K$ queries in the \texttt{dev} partition. The candidate passages are drawn from a corpus of over $8.8M$ passages. Each query is associated with relevant passages labeled with a relevance score of $1$, and irrelevant passages labeled with a score of $0$. Following the setup in RankT5~\citep{zhuang2022rankt5finetuningt5text}, our evaluation focuses on the top $1000$ retrieved documents using MRR@10 and NDCG@10 as metrics, while our training uses a sample of $36$ documents ($1$ positive plus sampled $35$ negatives) per query.

We compared the finetuned performance of Gemma Encoder against the RankT5 model~\citep{zhuang2022rankt5finetuningt5text}. As shown in Table~\ref{msmarco-results}, the Gemma 2B and 9B variants outperform RankT5, despite not leveraging any up-training or encoder-decoder MLM-style pretraining.

\begin{table} [h]
\small
\centering
\begin{tabular}{c c c c}
  \hline
  Model & Size & MRR@10 & NDCG@10\\ 
  \hline
  RankT5-XL & 1.5B & 0.4358 & 0.5035 \\ 
  Gemma-2 2B & 2B & \textbf{0.4456} & 0.5133 \\ 
  Gemma-2 9B & 9B & 0.4450 & \textbf{0.5148} \\ 
  \hline
\end{tabular}
\captionsetup{width=0.95\linewidth,font=footnotesize,justification=centering}
\caption{Evaluation on MS MARCO benchmark for Ranking tasks. \texttt{MRR} and \texttt{NDCG} stand for Mean Reciprocal Rank and Normalized Discounted Cumulative Gain, the standard ranking metrics. \label{msmarco-results}}
\end{table}

\begin{table}[h]
\small
\centering
\begin{tabular}{c c  c c c c}
  \hline
  \multirow{2}{*}{Model} & \multirow{2}{*}{First-K} & \multirow{2}{*}{Mean} & \multicolumn{2}{c}{Attention} & Last-K \\ 
  &&&\texttt{Q}&\texttt{KV}& \\
  \hline
  2B & 88.7 & \textbf{89.4} & 89.0 & 89.0 & 89.1 \\
  9B & 90.4 & 89.7 & 90.7 & 90.6 & \textbf{90.9} \\
  \hline
\end{tabular}
\captionsetup{width=0.95\linewidth,font=footnotesize,justification=centering}
\caption{Pooling strategy ablation. \texttt{Q} and \texttt{KV} denote Q-probe and KV-probe Attention Pooling, respectively. \textbf{Bidirectional} attention is used by default. \label{pooling-summary}}
\end{table}

\begin{table}[h]
\small
\centering
\begin{tabular}{c c c c c}
  \hline
  \multirow{2}{*}{Model} & \multirow{2}{*}{BiDi} & \multicolumn{3}{c} {Causal} \\
  && Mean & Attent & Last-K \\
  \hline
  2B & \textbf{89.4} & 84.5 & 86.0 & 88.6 \\ 
  9B & \textbf{90.9} & 87.5 & 88.4 & 90.4 \\ 
  \hline
\end{tabular}
\captionsetup{font=footnotesize,justification=centering}
\caption{Attention mechanism Ablation. \texttt{BiDi} and \texttt{Causal} stand for bidirectional and causal attention, respectively. The \texttt{BiDi} results are the best results from Table~\ref{pooling-summary}. For causal masking results, we applied \texttt{mean}, \texttt{attention} and \text{last token} poolings for comparison. Details in Sec.~\ref{sec:attention_ablation}. \label{attention-summary}}
\end{table}

\begin{table}[h]
\small
\centering
\begin{tabular}{c c c c c}
  \hline
  \multirow{2}{*}{Model} & \multicolumn{2}{c}{Bidirectional Attn.} & \multicolumn{2}{c} {Causal Attn.} \\
  & left & right & left & right \\
  \hline
  2B & 88.9 & \textbf{89.4} & 88.4 & 88.6 \\ 
  9B & 90.8 & \textbf{90.9} & 90.6 & 90.4 \\ 
  \hline
\end{tabular}
\captionsetup{width=0.95\linewidth,font=footnotesize,justification=centering}
\caption{Padding strategies ablation. No significant difference is found for comparing two choices of padding strategies after finetuning. \label{padding-summary}}
\end{table}

\begin{figure}[htp]
\centering
\includegraphics[scale=0.37]{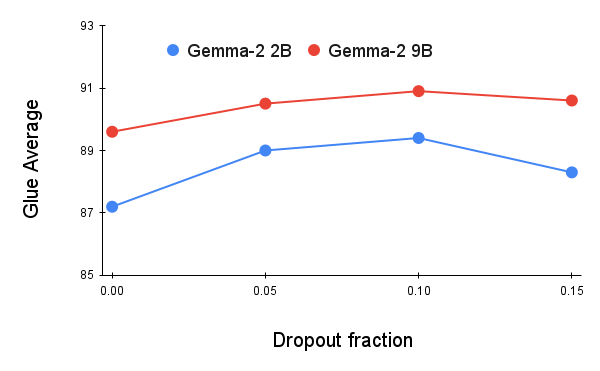}
\captionsetup{width=0.95\linewidth,font=footnotesize,justification=centering}
\caption{Effect of feed-forward and attention dropout. \label{dropout-summary}}
\end{figure}

\subsection{Ablation}
\label{sec:ablation}

To optimize performance, as demonstrated in Tables~\ref{glue-results} and~\ref{superglue-results}, we conducted ablation studies on architectural and hyperparameter choices.

\subsubsection{Pooling Strategy}
\label{sec:pooling_ablation}
To evaluate the influence of pooling strategy on the performance of the Gemma Encoder, we conducted experiments using the GLUE benchmark.  Following the approach in T5~\citep{raffel2020t5}, which demonstrated strong performance for encoder-based tasks, we utilized bidirectional attention masking during both finetuning and inference.

Table~\ref{pooling-summary} summarizes the GLUE average scores (refer to Table~\ref{tab:pooling-appendix} for per-task results) obtained with different pooling strategies for the Gemma-2 2B and 9B models~\citep{gemmateam2024gemma2improvingopen}.  Our analysis reveals that simple pooling strategies, specifically last-token and mean pooling, achieve superior performance compared to attention pooling. But with bidirectional attention, no single pooling strategy consistently optimized performance for both Gemma 2B and 9B. 

\textbf{Note on limitation}: It is important to acknowledge that the observed superiority of simple pooling strategies is specific to the GLUE benchmark finetuning context, which is characterized by relatively limited training data.  In scenarios involving large-dataset finetuning, or pre-finetuning a decoder, especially for retrieval tasks (e.g. \citet{lee2024geckoversatiletextembeddings} for pretrained retrieval model, \citet{moiseev2023samtoneimprovingcontrastiveloss, dong2022exploringdualencoderarchitectures} for finetuning), a reevaluation of the relative efficacy of attention pooling and simple pooling methods is warranted.  The increased parameter count associated with attention pooling is a significant factor to consider in such contexts. Please refer to Section~\ref{sec:app_attention_pooling} for more details. 

\subsubsection{Attention Masking}
\label{sec:attention_ablation}
To investigate the impact of attention masking strategies on Gemma Encoder performance, we evaluated both causal and bidirectional masking using the GLUE benchmark.  Table~\ref{attention-summary} provides a comparison of the GLUE average scores for Gemma-2 2B and 9B models under both masking conditions.  For the bidirectional masking results, we report the optimal performance achieved across different pooling strategies, consistent with the methodology in Table~\ref{pooling-summary}.

A key observation from our experiments is the superior performance of bidirectional masking compared to causal masking. This pattern is consistently observed across the majority of tasks within the GLUE benchmark, detailed in Section~\ref{sec:app_attention} for different model sizes. This finding is particularly noteworthy given that the Gemma decoder's pretraining process utilizes solely causal masking. It suggests that employing bidirectional masking, at finetuning time, can still enhance performance on encoder-related tasks, even with a causally pretrained decoder and limited finetuning data.

Furthermore, our results demonstrate that when causal masking is employed, last-token pooling exhibits significantly better performance than alternative pooling methods. This observation aligns with the inherent nature of causal masking, where the last token possesses a comprehensive contextual understanding due to its attention over the entire preceding sequence. This is analogous to the pretraining objective of decoder-based language models, where the embedding of the last token serves as the basis for predicting the subsequent token.

\subsubsection{Dropouts}
\label{sec:dropout_ablation}

Considering the task-specific nature and deterministic training objectives commonly associated with encoder tasks, we investigated the influence of dropout regularization on the performance of the Gemma Encoder using the GLUE benchmark. Figure~\ref{dropout-summary} presents the GLUE average scores for the Gemma-2 2B and 9B models as a function of increasing dropout rate, starting from a rate of zero.

Our analysis reveals that the application of dropout generally enhances the performance of the Gemma Encoder for both the 2B and 9B models.  However, a performance degradation is observed when the dropout rate exceeds 0.1.  This pattern is consistently observed across the majority of tasks within the GLUE benchmark, detailed in Section~\ref{sec:app_dropout} for both model sizes.  Therefore, based on these findings, we recommend a dropout rate of 0.1 applied to both the attention softmax and feedforward network outputs during the finetuning process for encoder-based tasks.

\subsubsection{Padding Strategy}
\label{sec:padding_ablation}
To evaluate the influence of padding side on the performance of the Gemma Encoder, we compared left and right padding strategies. Table~\ref{padding-summary} summarizes the GLUE average scores obtained for the Gemma-2 2B and 9B models under both padding conditions. Consistent with our expectations, no statistically significant difference in performance was observed between the two padding approaches. This indicates that the model exhibits robustness to padding side after finetuning.

\section{Conclusion}

We introduce Gemma Encoder, adapting the decoder-only Gemma model for encoder tasks. We addressed using decoder models' pre-trained knowledge for tasks typically handled by encoders (e.g., BERT, T5's encoder). Our approach involved task-specific pooling/MLP layers, bidirectional attention, and dropout during finetuning. We also analyzed padding strategies.

Gemma Encoder outperformed baselines on GLUE, SuperGLUE, and MSMARCO benchmarks (classification, regression, ranking), proving that well-adapted decoder-only models excel at encoder tasks. Bidirectional attention proved crucial for capturing context, and dropout improved robustness and generalization. Adaptable pooling and output layers further enhanced performance.

\newpage

\begin{table*}[h]
\fontsize{9.5pt}{9.5pt}\selectfont
\centering
\begin{tabular}{c c c c c c c c c c c c}
  \hline \hline
  \noalign{\vskip 1mm}    
    \multirow{2}{*}{Model} & \multirow{2}{*}{Pooling} & \multirow{2}{*}{STSB} & \multirow{2}{*}{CoLA} & \multirow{2}{*}{QQP} & \multirow{2}{*}{QNLI} & \multirow{2}{*}{SST2} & \multirow{2}{*}{RTE} & \multirow{2}{*}{MRPC} & \multicolumn{2}{c}{MNLI} &  \multirow{2}{*}{Avg}\\ 
  &&&&&&&&&m&mm& \\
  \hline
  \noalign{\vskip 1mm}    
      \multirow{4}{*}{Gemma 2B} & First-K & 92.3 & 65.5 & 88.3 & 95.2 & 96.7 & 87.0 & 92.2 & 90.6 & 90.9 & 88.7 \\
      & Mean & 92.4 & 67.7 & 89.2 & 95.4 & 97.0 & 87.3 & 93.3 & 91.1 & 91.1 & 89.4 \\
      & Attention & 92.3 & 67.1 & 88.7 & 95.3 & 96.7 & 87.3 & 92.0 & 90.7 & 90.9 & 89.0 \\
      & Last token & 92.4 & 65.7 & 88.1 & 95.2 & 97.4 & 88.8 & 92.7 & 90.8 & 90.8 & 89.1 \\
  \noalign{\vskip 1mm}
  \hline
  \noalign{\vskip 1mm}    
      \multirow{4}{*}{Gemma 9B} & First-K & 92.1 & 72.3 & 89.2 & 96.1 & 96.3 & 90.6 & 93.0 & 92.4 & 92.0 & 90.4 \\
      & Mean & 91.5 & 69.5 & 90.3 & 96.4 & 96.5 & 88.8 & 89.4 & 92.4 & 92.2 & 89.7 \\
      & Attention & 92.5 & 72.1 & 90.1 & 96.5 & 96.5 & 91.3 & 93.1 & 92.5 & 92.1 & 90.7 \\
      & Last token & 92.4 & 70.9 & 89.8 & 96.3 & 96.7 & 94.2 & 93.2 & 92.3 & 92.1 & 90.9 \\
  \noalign{\vskip 1mm}    
  \hline \hline
\end{tabular}
\captionsetup{width=0.95\textwidth,font=scriptsize,justification=centering}
\caption{Impact of pooling on GLUE task performance. Experiments used bidirectional attention, 10\% dropout, and right padding. Attention pooling utilizes a Query probe. \label{tab:pooling-appendix}}
\end{table*}

\begin{table*}[!h]
\fontsize{9.5pt}{9.5pt}\selectfont
\centering
\begin{tabular}{c c c c c c c c c c c c}
  \hline \hline
  \noalign{\vskip 1mm}    
    \multirow{2}{*}{Model} & \multirow{2}{*}{Attention} & \multirow{2}{*}{STSB} & \multirow{2}{*}{CoLA} & \multirow{2}{*}{QQP} & \multirow{2}{*}{QNLI} & \multirow{2}{*}{SST2} & \multirow{2}{*}{RTE} & \multirow{2}{*}{MRPC} & \multicolumn{2}{c}{MNLI} &  \multirow{2}{*}{Avg}\\ 
  &&&&&&&&&m&mm& \\
  \hline
  \noalign{\vskip 1mm}    
      \multirow{4}{*}{Gemma 2B} & Causal (Mean) & 90.7 & 65.2 & 87.7 & 94.4 & 96.2 & 61.0 & 85.4 & 89.8 & 90.1 & 84.5 \\
      & Causal (Attent.) & 91.0 & 68.3 & 87.9 & 94.7 & 95.8 & 69.3 & 86.1 & 90.5 & 90.7 & 86.0 \\
      & Causal (Last-K) & 92.1 & 67.3 & 88.1 & 94.9 & 96.3 & 86.6 & 90.6 & 90.6 & 90.7 & 88.6 \\
      & BiDi & 92.4 & 67.7 & 89.2 & 95.4 & 97.0 & 87.3 & 93.3 & 91.1 & 91.1 & 89.4 \\
  \noalign{\vskip 1mm} 
  \hline
  \noalign{\vskip 1mm} 
      \multirow{4}{*}{Gemma 9B} & Causal (Mean) & 91.7 & 72.5 & 89.4 & 95.9 & 96.3 & 73.6 & 85.0 & 91.8 & 91.7 & 87.5 \\
      & Causal (Attent.) & 92.1 & 71.9 & 89.6 & 96.1 & 96.5 & 76.8 & 88.3 & 92.0 & 91.9 & 88.4 \\
      & Causal (Last-K) & 92.4 & 72.2 & 89.4 & 95.9 & 96.5 & 91.7 & 91.8 & 91.9 & 91.8 & 90.4 \\
      & BiDi & 92.4 & 70.9 & 89.8 & 96.3 & 96.7 & 94.2 & 93.2 & 92.3 & 92.1 & 90.9 \\
  \noalign{\vskip 1mm}
  \hline \hline
\end{tabular}
\captionsetup{width=0.95\textwidth,font=scriptsize,justification=centering}
\caption{Impact of attention mechanism on GLUE tasks: \texttt{BiDi} (bidirectional) uses optimal pooling; \texttt{Causal} uses mean, attention, and last-token pooling. All experiments use 10\% dropout and right padding. Attention pooling utilizes a Query probe.\label{tab:attention-appendix}}
\end{table*}

\begin{table*}[!h]
\fontsize{9.5pt}{9.5pt}\selectfont
\centering
\begin{tabular}{c c c c c c c c c c c c}
  \hline \hline
  \noalign{\vskip 1mm}
 \multirow{2}{*}{Model} & \multirow{2}{*}{Dropout} & \multirow{2}{*}{STSB} & \multirow{2}{*}{CoLA} & \multirow{2}{*}{QQP} & \multirow{2}{*}{QNLI} & \multirow{2}{*}{SST2} & \multirow{2}{*}{RTE} & \multirow{2}{*}{MRPC} & \multicolumn{2}{c}{MNLI} &  \multirow{2}{*}{Avg}\\ 
  &&&&&&&&&m&mm& \\
  \noalign{\vskip 1mm}
  \hline
  \noalign{\vskip 1mm}
      \multirow{4}{*}{Gemma 2B} & 0 & 91.0 & 62.8 & 89.2 & 95.2 & 96.2 & 78.3 & 89.0 & 91.3 & 91.5 & 87.2 \\
      & 0.05 & 92.0 & 66.4 & 88.9 & 95.3 & 96.4 & 87.3 & 92.4 & 91.0 & 91.2 & 89.0 \\
      & 0.10 & 92.4 & 67.7 & 89.2 & 95.4 & 97.0 & 87.3 & 93.3 & 91.1 & 91.1 & 89.4 \\
      & 0.15 & 92.0 & 66.5 & 88.3 & 95.0 & 96.7 & 84.4 & 91.0 & 90.5 & 90.7 & 88.3 \\
  \noalign{\vskip 1mm}
  \hline
  \noalign{\vskip 1mm}
      \multirow{4}{*}{Gemma 9B} & 0 & 91.3 & 68.5 & 90.2 & 96.4 & 96.6 & 88.8 & 90.0 & 92.4 & 92.2 & 89.6 \\
      & 0.05 & 92.2 & 71.0 & 90.0 & 96.3 & 96.4 & 92.4 & 92.0 & 92.3 & 92.0 & 90.5 \\
      & 0.10 & 92.4 & 70.9 & 89.8 & 96.3 & 96.7 & 94.2 & 93.2 & 92.3 & 92.1 & 90.9 \\
      & 0.15 & 92.3 & 71.4 & 89.5 & 96.2 & 96.6 & 92.4 & 93.3 & 92.1 & 92.0 & 90.6 \\
  \noalign{\vskip 1mm}
  \hline \hline
\end{tabular}
\captionsetup{width=0.95\textwidth,font=scriptsize,justification=centering}
\caption{Impact of dropout on GLUE task performance. Experiments used bidirectional attention and right padding. Gemma 2B employed mean pooling, while Gemma 9B used last-token pooling. \label{tab:dropout-appendix}}
\end{table*}

\begin{table*}[!h]
\fontsize{9.5pt}{9.5pt}\selectfont
\centering
\begin{tabular}{c c c c c c c c c c c c c c}
  \hline \hline
  \noalign{\vskip 1mm}
 \multirow{2}{*}{Probe} & \multirow{2}{*}{H} & \multirow{2}{*}{T} & \multirow{2}{*}{\#params} & \multirow{2}{*}{STSB} & \multirow{2}{*}{CoLA} & \multirow{2}{*}{QQP} & \multirow{2}{*}{QNLI} & \multirow{2}{*}{SST2} & \multirow{2}{*}{RTE} & \multirow{2}{*}{MRPC} & \multicolumn{2}{c}{MNLI} &  \multirow{2}{*}{Avg}\\ 
  &&&&&&&&&&&m&mm& \\
  \noalign{\vskip 1mm}
  \hline
  \noalign{\vskip 1mm}
      Q & 1 & 1 & 2.3M & 92.3	& 67.1	& 88.7	& 95.3	& 96.7	& 87.3	& 92.0	& 90.7	& 90.9	& 89.0 \\
      Q & 2 & 1 & 4.7M & 92.2	& 66.4 & 88.6 & 95.1 & 96.6 & 86.3 & 92.1 & 90.6 & 90.6 & 88.7 \\
      Q & 4 & 1 & 9.4M & 92.2	& 67.4 & 88.8 & 95.1 & 96.7 & 86.6 & 92.2 & 90.6 & 90.6 & 88.9\\
      Q & 8 & 1 & 18.8M & 92.2	& 66.4 & 88.6 & 95.4 & 96.8 & 85.6 & 91.9 & 90.7 & 90.7 & 88.7 \\
      Q & 1 & 512 & 2.3M & 92.0 & 66.1 & 88.7 & 95.3 & 96.8 & 85.9 & 91.7 & 90.6 & 90.6 & 88.6 \\
      Q & 8 & 512 & 18.8M & 92.2 & 66.1 & 88.7 & 95.3 & 96.8 & 84.1 & 91.9 & 90.5 & 90.5 & 88.5 \\
  \noalign{\vskip 1mm}
  \hline
  \noalign{\vskip 1mm}
      KV & 1 & 512 & 2.3M & 92.3 & 66.6 & 88.5 & 95.1 & 96.9 & 88.4 & 92.2 & 90.6 & 90.6 & 89.0 \\
      KV & 8 & 512 & 18.8M & 92.3 & 67.9 & 88.6 & 95.1 & 96.7 & 87.0 & 92.2 & 90.7 & 90.7 & 89.0 \\
  \noalign{\vskip 1mm}
  \hline \hline
\end{tabular}
\captionsetup{width=0.95\textwidth,font=scriptsize,justification=centering}
\caption{Gemma 2B performance on GLUE tasks was evaluated with ablations on \textbf{attention pooling probes}, varying the number of attention heads (\textbf{H}) and pooled tokens (\textbf{T}). Experiments used bidirectional attention, 10\% dropout, and right padding. \label{tab:attention-pooling-appendix}}
\end{table*}

\appendix
\section{Additional Results}
\subsection{Pooling Ablation Details}
\label{sec:app_pooling}
Table~\ref{tab:pooling-appendix} provides a breakdown of per-task performance on the GLUE benchmark for different pooling strategies.  The query-probe approach is utilized for attention pooling.  The results indicate that simple pooling strategies achieve performance competitively to that of attention pooling, even though attention pooling introduces a greater number of parameters.

\subsection{Attention Ablation Details}
\label{sec:app_attention}

Table~\ref{tab:attention-appendix} provides a breakdown of per-task performance on the GLUE benchmark for different attention mechanisms.

\subsection{Dropout Ablation Details}
\label{sec:app_dropout}

Table~\ref{tab:dropout-appendix} provides a breakdown of per-task performance on the GLUE benchmark across a range of dropout rates.

\subsection{Attention Pooling Ablation}
\label{sec:app_attention_pooling}

Table~\ref{tab:attention-pooling-appendix} provides a comparison of per-task performance on the GLUE benchmark for various attention pooling variants.  An increase in attention complexity, reflected in a larger number of parameters, resulted in a decrease in performance on the GLUE benchmarks.  As detailed in Table~\ref{data-stats}, the GLUE training sets are limited in size (all containing fewer than 1 million examples), which is inadequate for the effective adaptation of a randomly initialized attention pooler with millions of parameters.

\vskip 0.2in
\bibliography{references}

\end{document}